A THESIS ON

# "SIX DEGREE ROBOTIC ARM WITH MIMICKING MECHANISM"

**By:-**
**PARAM KOTHARI**

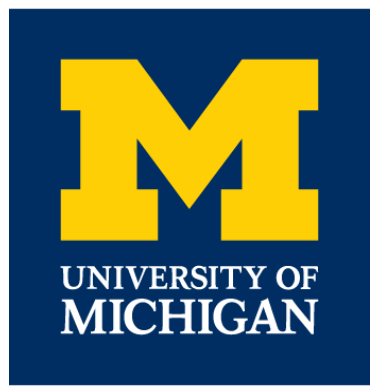

**College of Engineering**

**Global Automotive and Manufacturing Engineering**

**Integrative Systems + Design**

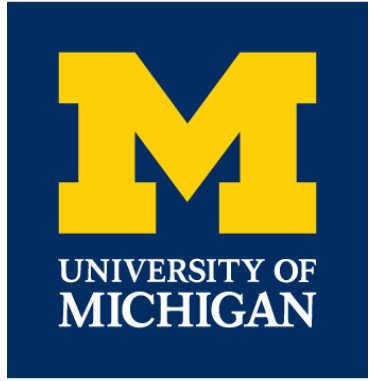

# DECLARATION OF ORIGINALITY PAGE

I certify that, to the best of my knowledge, my thesis does not infringe upon anyone's copyright nor violate any proprietary rights and that any ideas, techniques, quotations, or any other material from the work of other people included in my thesis, published or otherwise, are fully acknowledged in accordance with the standard referencing practices. Furthermore, to the extent that I have included copyrighted material that surpasses the bounds of fair dealing within the meaning of the Copyright Law of the United States, I certify that I have obtained a written permission from the copyright owner(s) to include such material(s) in my thesis and have included copies of such copyright clearances to my appendix.

I declare that this is a true copy of my thesis, including any final revisions.

**Name of Student:**         **PARAM KOTHARI**

# ACKNOWLEDGMENT

Foremost, I would like to express my sincere gratitude to professor **Mr. Tejas Raval (Asst. Prof. Automobile Department, IITE)** for the continuous support of my Project study and research, for his patience, motivation, enthusiasm, and immense knowledge. His guidance helped me in all the time of research and writing of this report. I could not have imagined having a better advisor and mentor for my Project study.

Moreover, I would also like to thank **Mr. Omkar Bhatt,** CEO & Founder of **OMNI Automation** and **Mr. Jay Mehta**, a charismatic employee at **MC Machines** for their valuable guidance and support.

I offer my special gratitude to all the **faculty members** for their help and support. I thank my **friends** for providing me such a warm atmosphere to make my study more delightful and memorable.

I would like to express endless gratitude to "**My Parents**" who gave me everything they could to enable me to reach the highest possible education level. I only hope that they know how their love, support and patience encouraged me to fulfil my dream.

I would like to thank all people who have helped and inspired me during my report study.

# ABSTRACT

Multi-degree of freedom robots are playing very important role in different applications of automation. They are providing much more accuracy in carrying out a typical procedure as compared to the manual work done by human. In recent years the design, fabrication and development of robotic arms have been active research areas in robotics all around the world. This project describe a mechanical system, design concept and prototype implementation of a 6 DOF robotic arm, which should perform industrial task such as pick and place of fragile objects operation. This robot arm being controlled by micro-controller has base, shoulder, elbow, wrist rotation and a functional gripper. Gripper has been built as end-effector and is capable of grasping diverse objects within own workspace of the arm possible.

I made an effort to endeavour one of the best model possible out of the three concepts tried. The three concepts that I undertook were dependent on the type of mechanism used to provide the movement.

The first concept was to provide the movement with the help of Pneumatics. The idea of pneumatics did provide a feasible movement but I had to overcome the barrier of the weight as the weight of either the storage of gas or the air compressor would have affected the usage evidently. Similarly, in the second concept, I dealt with the forces of Hydraulics. The forces exerted by hydraulics, not only restricted the movement, but also annexed a certain amount of delay in the time period in which the movement had to be done. In, the third concept, I tried to provide the necessary movement with the help of Servo Motors. The results achieved with servo motors in replacement of pneumatics and hydraulics were impeccable. The motors not only provided the movement with minimal lag but also did not impede the movement of the hand.

# CONTENT

# LIST OF FIGURE

# LIST OF SYMBOLS, ABBREVATIONS & ACRONYMS

| SYMBOL | DESCRIPTION |
|---|---|
| N | Degree Of Freedom |
| C | Configuration Space |
| Dim(C) | Dimension of 'C' |
| c | Radius of small circular ring |
| X | 1st Axis in the Plane |
| Y | 2nd Axis in the Plane |
| x | Cartesian Coordinate |
| y | Cartesian Coordinate |
| r | Planner Polar Coordinate |
| φ | Planner Polar Coordinate |
| $n_c$ | Number of Independent Holonomic Constraints |
| $q_j$ | Configuration Variables |
| g | Grams |
| mm | Millimetre |
| m/s | Meters per Second |
| $m/s^2$ | Meters per Second squared |

| ABBREVATIONS | FULL-FORMS |
|---|---|
| Sin | Sine |
| **Cos** | Cosine |
| **Abd-Add** | Abduction-Adduction |
| **Flx-Ext** | Flexion-Extension |
| **Rad-Uln** | Radial-Ulnar |

| **ACRONYMS** | **FULL-FORMS** |
|---|---|
| **6 DOF** | Six Degrees Of Freedom |
| **3D** | Three Dimensional |
| **MPU** | Magnetic Pickup |
| **GY** | Gyroscope |
| **DMP** | Digital Motion Processor |
| **MEMS** | Micro electrical Mechanical Systems |
| **MMI** | Man Machine Interface |
| **DH** | Denvait Hartenberg |
| **I$^2$C** | I-squared-C |
| **VCC** | Voltage Common Collector |
| **PWM** | Pulse Width Modulator |
| **PCB** | Printed Circuit Board |
| **IDE** | Integrated Development Environment |

# Chapter 1

# INTRODUCTION

In India, the physically challenged people consists about 2.21% of the whole population. This accounts to nearly 2.68 crore people. A number of firms, non-governmental organizations, trust funds work for the welfare of such people. I also wanted to contribute, but not in the form of money or service. And this lead us to thinking 'How can we empower them?' and further 'How about an artificial arm that functions better than the normal inborn human arm? '

The increment of the need for such products and that too at a lower cost has increased moderately. There were a gazillion number of ways to accomplish such a task, so I started off with a simple idea of making a simple mechanical arm that would mimic the movements of another arm on which the sensor would have been placed. This was just a stepping stone and would have paved our path to the ultimate goal of creating an artificial limb.

Moreover, the prototype can serve as an aid in different industries such as the space industry, medical industry and automobile manufacturing industry. To use it in the space industry, many changes would be necessary along with replacement of the materials with space grade aluminium. But after implementation and usage of such bots spacewalks would be considered redundant to a certain degree. Seeing the prototype's prominent usage in the medical industry to perform surgeries without the doctor's presence in the operation theatre, it would be safe to say that this kind of robotic arms have a promising future.

Several studies and projects made in the past have tried to develop a similar functioning robotic arm, but were not up to the mark. The precision and accuracy provided by the robotic arms of the past has been significantly low and the arms have been bulky too. The earliest version of this were developed by Johns Hopkins applied physics laboratory in United States of America which was a project of worth 120 million dollars of funding from the Defence Advanced Research Project Agency in the year 2015.

Technologies have shown great advancement and the bionic product industry has burgeoned prominently in the recent years. But the main stumbling block in the present market of bionics is the immensely high cost of the product. Many companies are trying to induce numerous products (along with different attachments) into the market as to acquire majority of the market with a lower rate.

More responsive robotic limbs will be created in future. With the flourishing 3D printing industry, advancements might help to reduce the cost of production and can even improve the aesthetic look of the artificial limb. With the advance processors, the limb might not provide any performance lag at all.

# Chapter 2

# THEORETICAL ASPECT

In order to put the analysis of a Robotic Arm, certain calculations must be made and basic forces must be configured before making a prototype or an actual structure. All the components of the forces must be considered heedfully at the time of designing the model. For now, Let us start with the basics.

## 2.1 What is a Robotic Arm ?

A Robotic Arm is very different from a Robotic Hand. A Robotic Arm is a type of mechanical arm, usually either programmable or computable. The arm maybe a quantum of the mechanism or maybe a part of a more complex robot. The linkages of such a mechanism are often affixed by joints allowing either rotational motion or translational displacement. The linkages of the mechanism can be considered to form a kinematic chain whose terminus can act as end effector that can seem to be analogous to the human arm to a certain extent.

A Robotic Arm is usually categorized on the basis of the coordinate system the arm uses to specify the location of the object as well as its own. Some of the Robots based on the above reasoning are :

- Articulated Robot
- Anthropomorphic Robot
- Cartesian Robot / Gantry Robot
- Cylindrical Robot

- Parallel Robot

- Polar Robot / Spherical Robot

- Selective Compliance Assembly Robot Arm / Selective Compliance Articulated Robot Arm

**2.2 What is Degree Of Freedom ( D O F ) ?**

Degree Of Freedom a.k.a. DOF, is a term of physics which helps to define the configuration state of a mechanical system, using the number of independent parameters associated with its manoeuvre. It is important in the analysis of mechanical system of bodies in the field of robotics.

The degrees-of-freedom of a mechanical system (denoted by N) may or may not equal the dimension of C (denoted by dim(C)). Consider, e.g., a particle free to move in the XY plane. Clearly, the particle has two degrees-of-freedom, namely: the two independent translations in the plane. These can be completely described by the Cartesian coordinates (x, y), or the planar polar coordinates (r, φ), where:

$x = r \cos \varphi$

$y = r \sin \varphi$

For such an unconstrained system, it is obvious that N = dim(C). However, mechanisms are typically constrained mechanical systems, and as such for them, the above equation takes the following form:

$N = \dim(C) - n_c$ ,

Where $n_c$ is the number of independent holonomic constraints. These constraints can be represented as equations in the configuration variables $q_j$ :

$$\eta_i(q_j) = 0, \quad i = 1, \ldots, n_c, \quad j = 1, \ldots, \dim(C).$$

The above can be compacted into a vector equation:

$$\eta(q) = 0$$

It is also important to note, that at least in some cases, it is possible to describe the same system as both constrained as unconstrained. For example, consider a small bead confined to a circular ring of radius c. The configuration space, C, in this case is trivially visualized as the circle. Since the circle is a planar entity, and the geometry of the circle is independent of the choice of its origin, path of the particle can be modelled as:

$$x^2 + y^2 - c^2 = 0$$

In the above description admits two variables: (x, y), and hence $\dim(C) = 2$. However, the particle has only one degree-of-freedom, which is along the ring itself. Therefore, in this case, $n_c = 2-1 = 1$, and Eq. (5) gives the functional form of the only constraint applicable.

The use of polar coordinates in the above case obviates the need for the constraint equation in the explicit form as above. In this case, the constraint can be absorbed in the description: r = c. Therefore, the remaining variable φ becomes free, and therefore in this description of the problem, $N = \dim(C) = 1$.

### 2.2.1 Six Degrees Of Freedom ( 6 D O F )

Six degrees of freedom (6 D O F) refers to the freedom of movement of a rigid body in three dimensional (3 D) space. Specifically, the body is free to move forward / backward, up / down, left / right (translation in three perpendicular axes) combined with rotation about three perpendicular axes, denominated as pitch, yaw, and roll.

This number six refers to the number of single-axis rotational joints in the arm, where the number indicates an increased flexibility in positioning a tool. This is a practical metric, in contrast to the abstract definition of degrees of freedom which measures the aggregate positioning capability of a system.

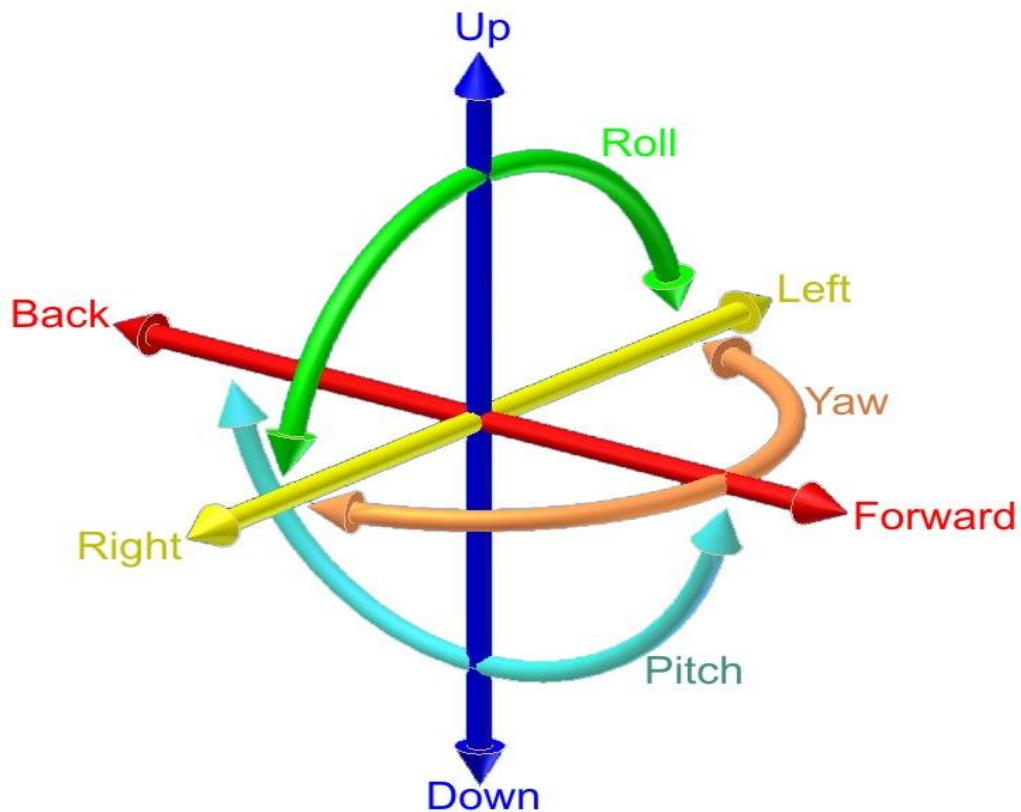

**Fig 2.1 : Six Degrees Of Freedom (6 D O F)**

## 2.3 What is a Mimicking Mechanism ?

To mimic someone / something is to imitate someone / something. I have attempted to develop a mechanism that helps artificial arm to replicate the actions and motions of the arm on which sensors are placed. The controller arm will be wearing a glove / sleeve with sensors placed on it, and will sense the motion of the joints of the arm which will be considered as the input.

### 2.3.1 MPU 6050

The MPU 6050 is an IMU with a MEMS accelerometer & gyroscope, and contains a 16-bit ADC. It has 6-DOF , i.e. x, y, & z for accelerometer and roll, pitch & yaw for gyroscope. In the project, I have used GY – 521, which is a breakout board for the MPU 6050 which extracts raw accelerometer and gyroscope data from the MPU.

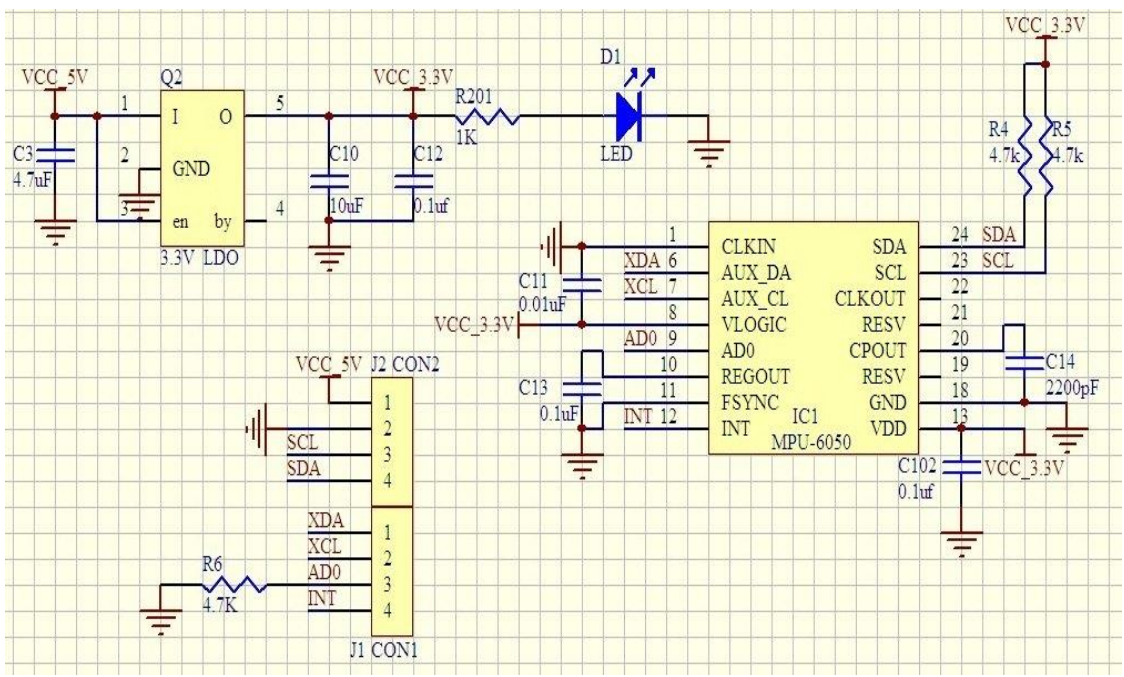

**Fig 2.2 : MPU 6050 GY – 521 Accelerometer Gyro-meter**

## 2.4 How is it different than a Normal Robotic Arm and Better than an inborn Natural Human Arm?

A Normal Robotic Arm can vary anywhere from 2 degrees of freedom to 20 degrees of freedom. The developed Robotic Arm consists of 6 Degrees Of Freedom.

The human arm is undeniably work of nature. But this version of hand is a result of millions of years of evolution and adaptation. I have designed a robotic arm that gives ability to perform actions of a human arm which will be mutually controlled by another human arm at a distance. In simple words this mechanical arm will always copy human arm's movements. In order to achieve this goal, I needed some equipment to convert human arm's mechanical motion into electrical signal so that that signal can be transferred to the power arm via communication channel. That equipment is the sensor on control arm which senses the arm's movement.

This project is actually made on the principles of an actual human arm. Rotation of the shoulder, lifting action of the elbow and grasping of the fingers are the most important movements that are mimicked.

In order to achieve this goal, we need some equipment to convert human arms mechanical motion into electrical signal so that that signal can be transferred to power arm via communication channel. That equipment is the sensor or control arm which sense the arm's movement. In this project, control arm or sensor arm acts as a master controller in a master-slave relation where power arm behaves like slave. It is a wearable assembly that consists of sensors which measures the change in angle of the motion of human arm and hand.

# Chapter 3

# LITERATURE SURVEY

## 3.1 Literature Survey

A considerable amount of investigations has been directed towards the prediction and measurement of forces acting on the Robotic Arm. That is because the forces generated during actions like grasping, lifting, moving the arm, etcetera, have a direct influence on the angle of the servo motor and accuracy of the. Due to the complex circuits and logic of the working of the Robotic Arm and some unknown factors and stresses, theoretical force were difficult to be calculated. Therefore, I had to take help from a number of research papers.

Some of the research papers focused on manufacturing aspect of the robotic arm while some focused on the force calculation aspect of the arm. I had to alter some of the calculation as they were calculated for different degrees of freedom.

### 3.1.1 The Concept Of Degrees Of Freedom.

Degree of freedom of a general mechanical system is defined as the minimum number of independent variables required to describe its configuration completely. The set of variables (dependent or independent) used to describe a system are termed as the configuration variables. For a mechanism, these can be either Cartesian coordinates of certain points on the mechanism, or the joint angles of the links, or a combination of both.

## 3.1.2 Design and Development of 6-DOF Robotic Arm Controlled by Man Machine Interface

This paper presents the design and development of a low cost and user friendly interface for the control of a 6-DOF slave tele-operated anthropomorphic robotic arm. Articulation of the robotic arm is achieved about six single-axis revolute joints: one for each shoulder abd-add, shoulder flx-ext, elbow flx-ext, wrist flx-ext, wrist rad-uln, and gripper open-close. Tele-operator, master, uses the Man Machine Interface (MMI) to operate in real-time the robotic arm. The MMI has simple motion capture devices that translate motion into analogue voltages which bring about the corresponding actuating signals in the robotic arm.

## 3.1.3 Modelling and Analysis of 6-DOF Robotic Arm Manipulator

The behaviour of physical systems in many situations may better be expressed with an analytical model. Robot modelling and analysis essentially involve its kinematics. For robotic manipulators having high Degrees Of Freedom (DOF) with multiple degrees in one or more joints, an analytical solution to the inverse kinematics is probably the most important topic in robot modelling. This paper develops the kinematic models a 6 DOF robotic arm and analyses its workspace. The proposed model makes it possible to control the manipulator to achieve any reachable position and orientation in an unstructured environment. The forward kinematic model is predicated on DH parametric scheme of robot arm position placement. Given the desired position and orientation of the robot end-effector, the realized inverse kinematics model provides the required corresponding joint angles. The forward kinematic model has been validated using Robotics Toolbox for MATLAB while the inverse kinematic model has been implemented on a real robotic arm. Experimental results demonstrate that using the developed model, the end-effector of robotic arm can point to the desired coordinates within precision of ±0.5cm. The approach presented in this work can also be applicable to solve the kinematics problem of other similar kinds of robot manipulators.

### 3.1.4 Dynamic Simulation and Motion Load Analysis of 6-DOF Robotic Articulated Robotic Arm

This research paper presents behaviour and performance of a six degree of freedom (DOF) robotic arm by using simulation through Autodesk Inventor. A 3D model of the manipulator was built; with a total workspace of 690 mm to carry a maximum payload of 200 g. Dynamic simulations are carried out on the model to examine the properties such as torque variation at the joints, speed, and trajectory of the manipulator. The simulation result showed that the manipulator has expected responses similar to a physical model. Maximum velocity of end effector is 0.6 m/s and acceleration of 0.59 m/s$^2$ . In the stress analysis, maximum Von-Mises stress was seen in the base which is equal to 14.91 MPa. This simulation method will reduce the need for testing on physical systems as long as the model sufficiently represents robotic arm to be built.

### 3.1.5 Kinematics Analysis & Modelling of a 6-DOF Robotic Arm

The aim of this study is to analyse the robot arm kinematics which is very important for the movement of all robotic joints. Also they are very important to obtain the indication for controlling or moving of the robot arm in the workspace. In this study the kinematics of Robotic Arm was accomplished by using LabVIEW. Finding the parameters of DH representation, the kinematic equations of motion can be derived which solve the problems of automatic control of the 6 revolute joints DFROBOT manipulator. The kinematics solution of the LabVIEW program was found to be nearest to the robot arms actual measurements

### 3.1.6 Dynamics and Motion of a 6-DOF Robot Manipulator

In this study a simple and direct solution to the mathematical model and kinematical analysis of the DFROBOT equations which relate all joints together as refer to the base is achieved. Applying the robot arm kinematics on LabVIEW, the manipulator motion can be introduced with respect to its mathematical analysis. In this study the target will be on the Kinematics and how to obtain the joint angles from the inverse kinematics modelling that can be used for the control of a variety of industrial processes. The work takes the benefit of using the numeric values for the position and orientation of the end effector which is the results of the forward kinematics and find all the joint angles of the robot arm from the inverse kinematic solutions applied in the new developed closed form package of the 6 DOF robot which is the case study of this study.

In this study a simple and direct solution to the mathematical model and kinematical analysis of the DFROBOT equations which relate all joints together as refer to the base is achieved. Applying the robot arm kinematics on LabVIEW, the manipulator motion can be introduced with respect to its mathematical analysis. In this study the target will be on the Kinematics and how to obtain the joint angles from the inverse kinematics modelling that can be used for the control of a variety of industrial processes. The work takes the benefit of using the numeric values for the position and orientation of the end-effector which is the results of the forward kinematics and find all the joint angles of the robot arm from the inverse kinematic solutions applied in the new developed closed form package of the 6 DOF robot which is the case study of this study.

### 3.1.7 Geometric Approach for Robotic Arm Kinematics with Hardware Design and Implementation

This paper presents a geometric approach to solve the unknown joint angles required for the autonomous positioning of a robotic arm. A plethora of complex mathematical processes is reduced using basic trigonometric in the modelling of the robotic arm. This modelling and analysis approach is tested using a five-degree-of-freedom arm with a gripper style end effector mounted to an iRobot Create mobile platform. The geometric method is easily modifiable for similar robotic system architectures and provides the capability of local autonomy to a system which is very difficult to manually control.

This paper aims to create a straightforward and repeatable process to solve the problem of robotic arm positioning for local autonomy. There have been many methods presented to allow this functionality. However, the majority of these methods use incredibly complex mathematical procedures to achieve the goals. Using a few basic assumptions regarding the working environment of the robot and the type of manipulation to take place, this paper proposes an easier solution which relies solely on the designation of a point in the three-dimensional space within the physical reach of the robotic arm. This solution has been achieved using a strictly trigonometric analysis in relation to a geometric representation of an arm mounted to a mobile robot platform.

The approach in this paper is similar to that used by Xu et al's 1000 Trials: An empirically validated end effector that robustly grasps objects from the floor in which an end effector is capable of retrieving various objects from the floor. The robot is assumed to have already located an object through various means and positioned itself in the correct orientation in front of the object. This robust grasping algorithm can then be combined with other work involving path planning, obstacle avoidance, and object tracking in order to produce a more capable robot.

# Chapter 4

# DESIGN AND FABRICATION

This chapter explains the steps that were taken in designing the parts for 3-D printing and how the electrical circuit was designed. It also explains the factors that were considered while designing. The design of the parts to be 3D printed was done in AutoCAD and CREO. The detailed design used all information from real world analysis, packing all the components in their specific location. The use of AutoCAD was done to finalise the design of the all the parts which will make the 3D printing process simplified. The program gives the dimension to all the parts including, bend angles, chamfer angles and Servo Motors parameters. This project was divided into several sections based on different areas of working.

**4.1 3D Printed Parts and Designing Logic**

All the parts used in making of this Robotic Arm are 3D printed. All the parts have been designed in such a way that the parts can easily be 3D printed with parameters as follow :

- 400 Micron Layer Height

- 3D Honeycomb Infill

- 35 % Infill for the parts.

Parts were printed with 3D Honeycomb Infill as it would have provided greater strength in all directions as compared to the rectangular fill pattern.

There are 3 sensors measuring the acceleration, one on the upper arm, second on the lower arm and third on the wrist. The hall sensor at the thumb just controls the gripper. Gravity is an acceleration pointing downwards. Reading the acceleration vectors, we can calculate the angles between the limbs and the gravity component. We use this to set the angles of the joints since the joints are relative to the ones up the kinematic chain, we have to calculate the relative angles. The relative elbow pitch is easily calculated by subtracting the shoulder pitch. The shoulder is the first joint here, so we can simply take the measured value. The wrist is the most complicated one since we have to measure a few vectors which are even affected by the lower arm roll. The yaw of the arm can only be measured using a compass, since it is perpendicular to the gravity, it has to be measured at the wrist to get the best deflection. But the responsible joint on the robot arm sits at the shoulder, so we have to subtract out any transformation happened in between and the function 'calculate angles' of the code calculates all angles and the transformation are done using some matrix multiplications.

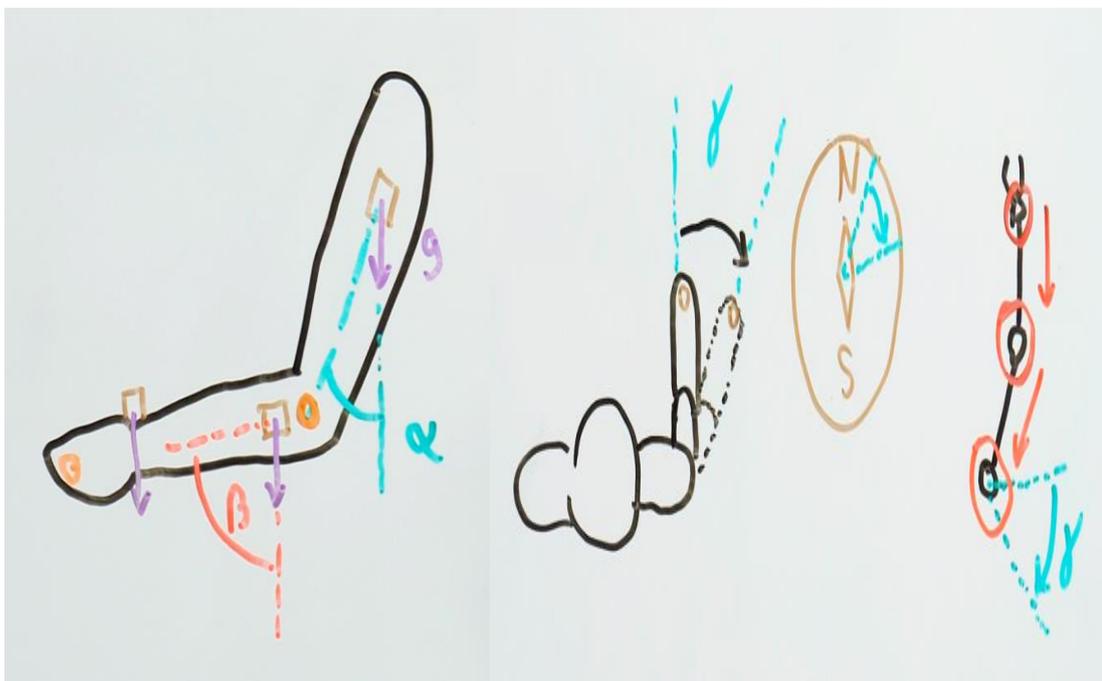

**Fig 4.1 : Forces acting on the Arm**

### 4.1.1 Base Servo Mount

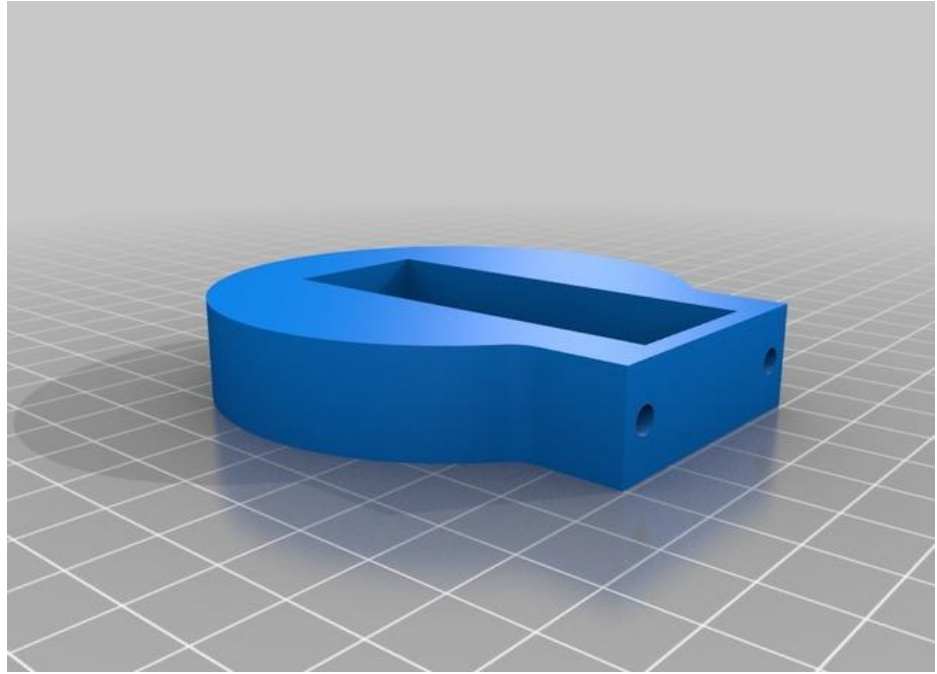

**Fig : 4.2**

### 4.1.2 Bearing Holder

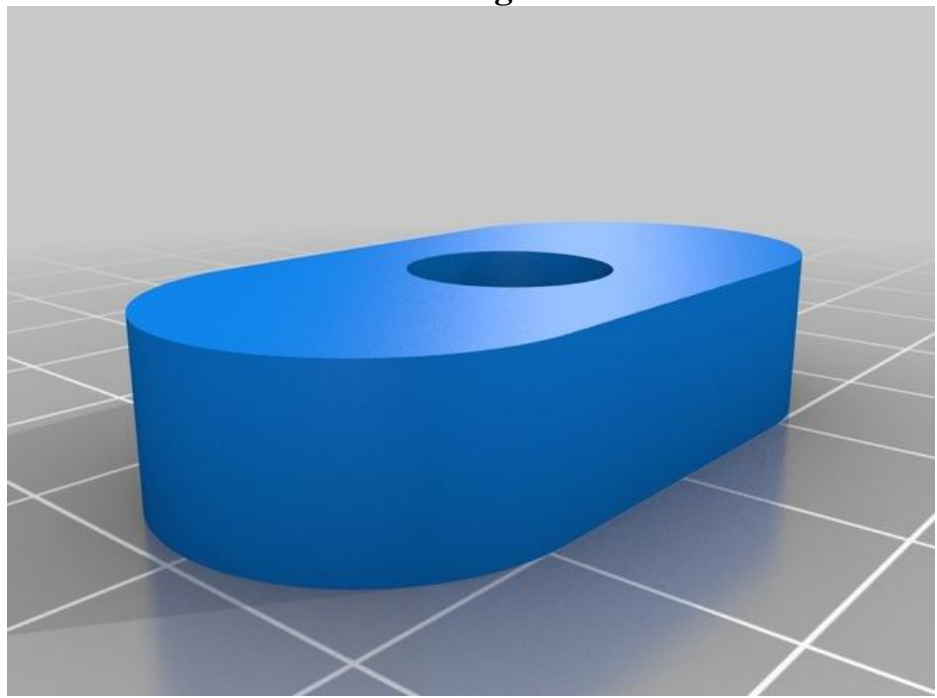

**Fig : 4.3**

### 4.1.3 Centre Arm Bearing Guide Clamp

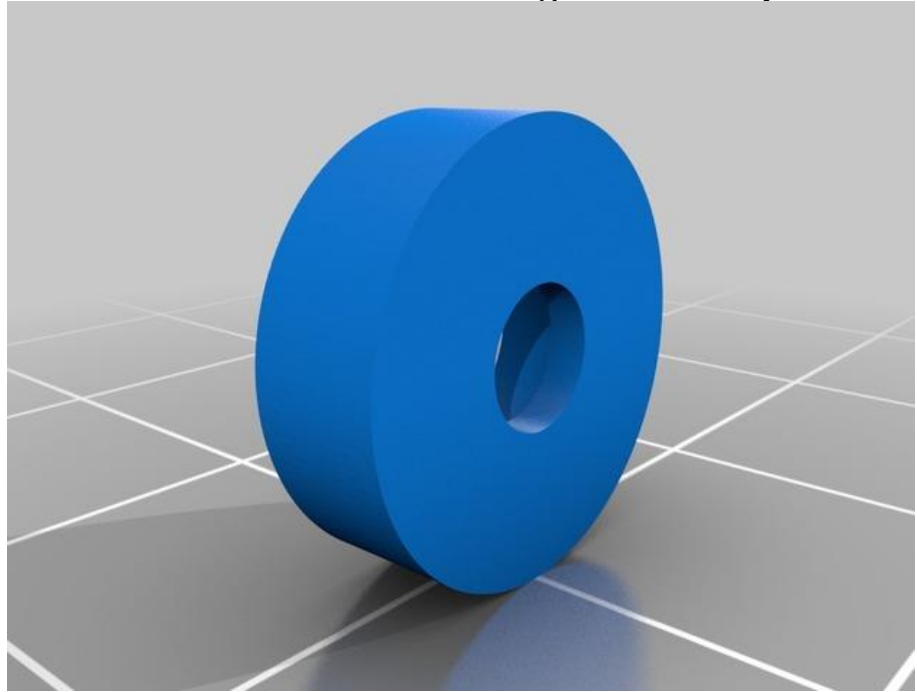

**Fig : 4.4**

### 4.1.4 Centre Arm Bearing Guide

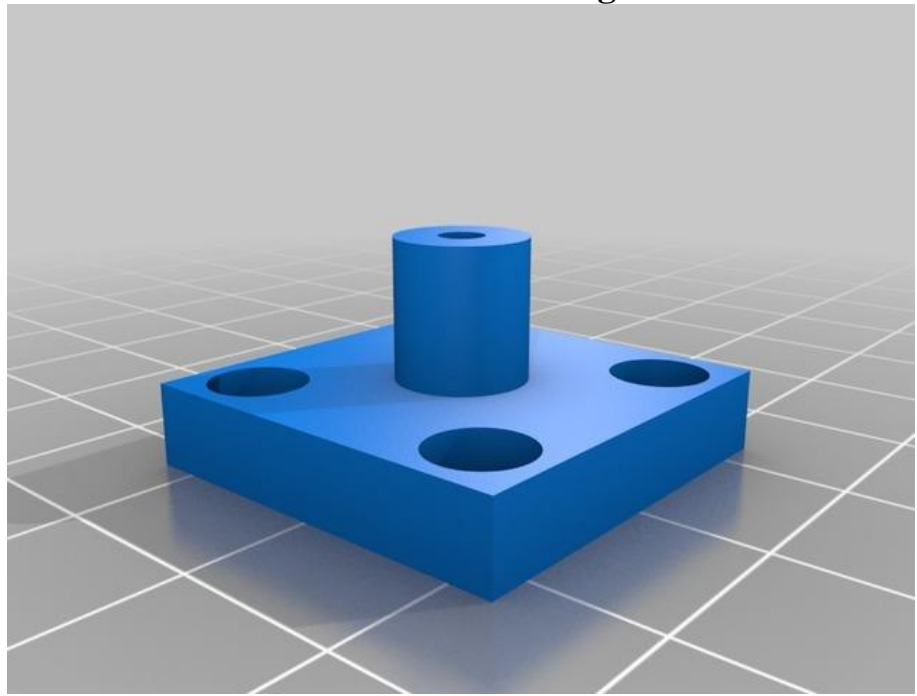

**Fig : 4.5**

### 4.1.5 Centre Arm

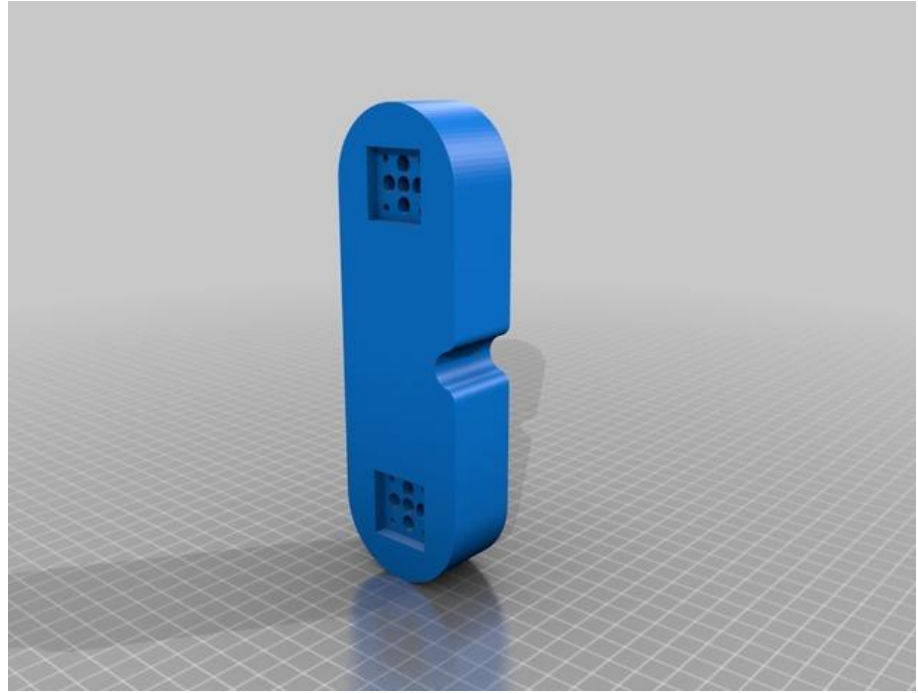

**Fig : 4.6**

### 4.1.6 Lower Arm Bushing

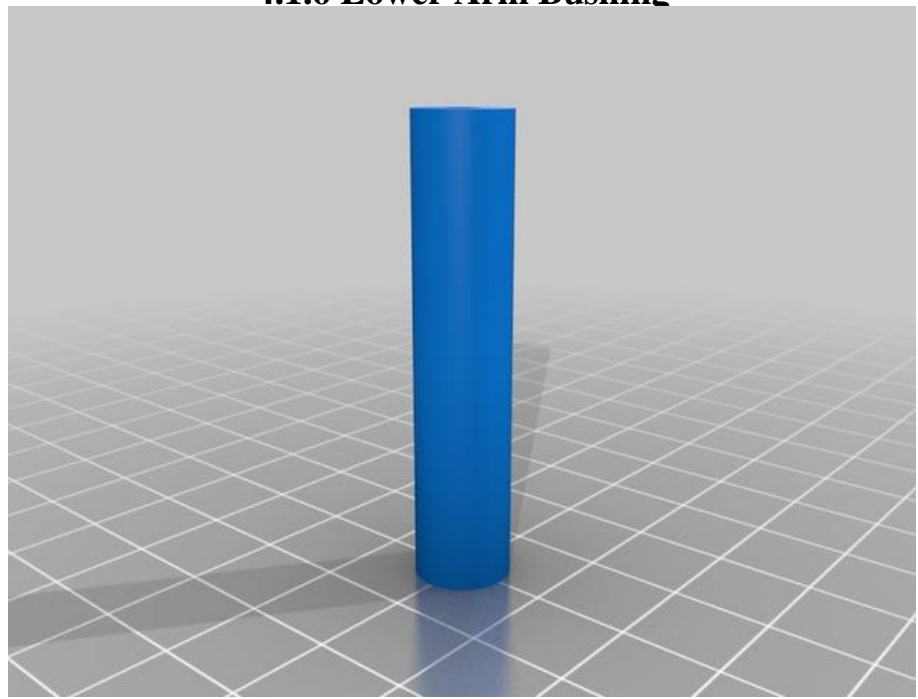

**Fig : 4.7**

### 4.1.7 Lower Arm Left

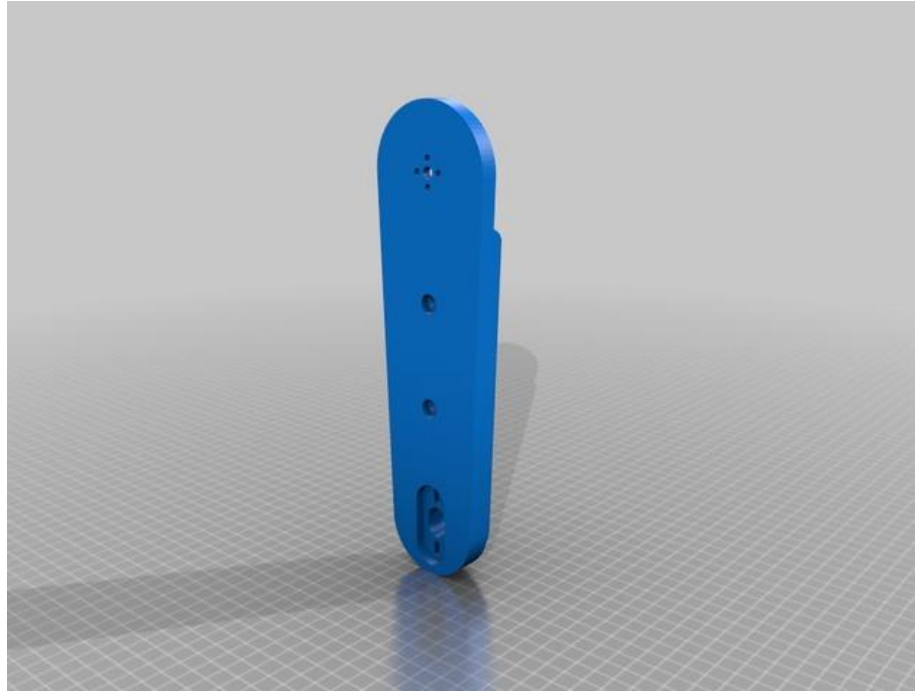

**Fig : 4.8**

### 4.1.8 Lower Arm Right

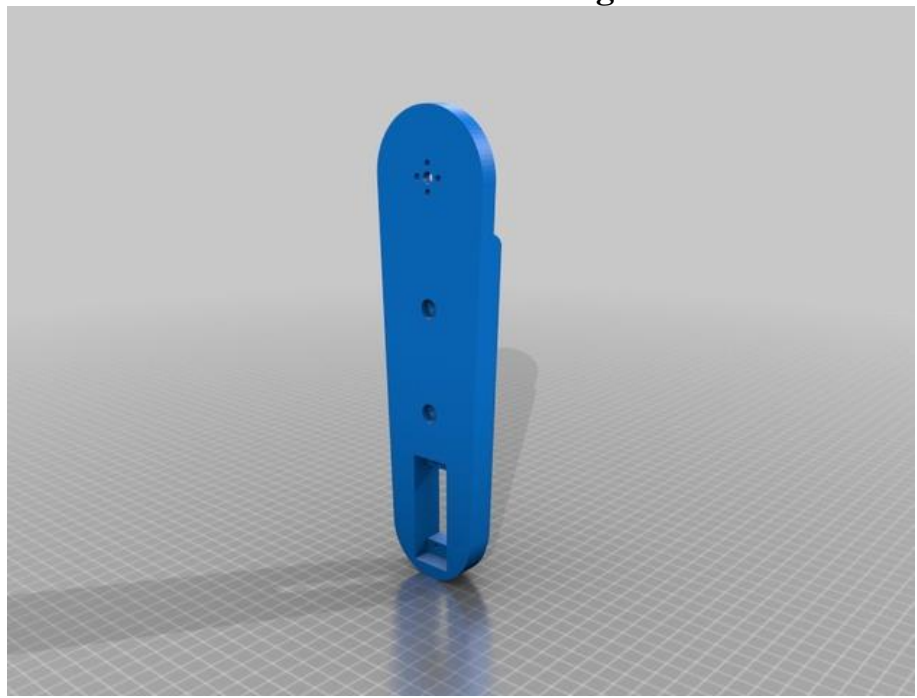

**Fig : 4.9**

### 4.1.9 Upper Arm Left

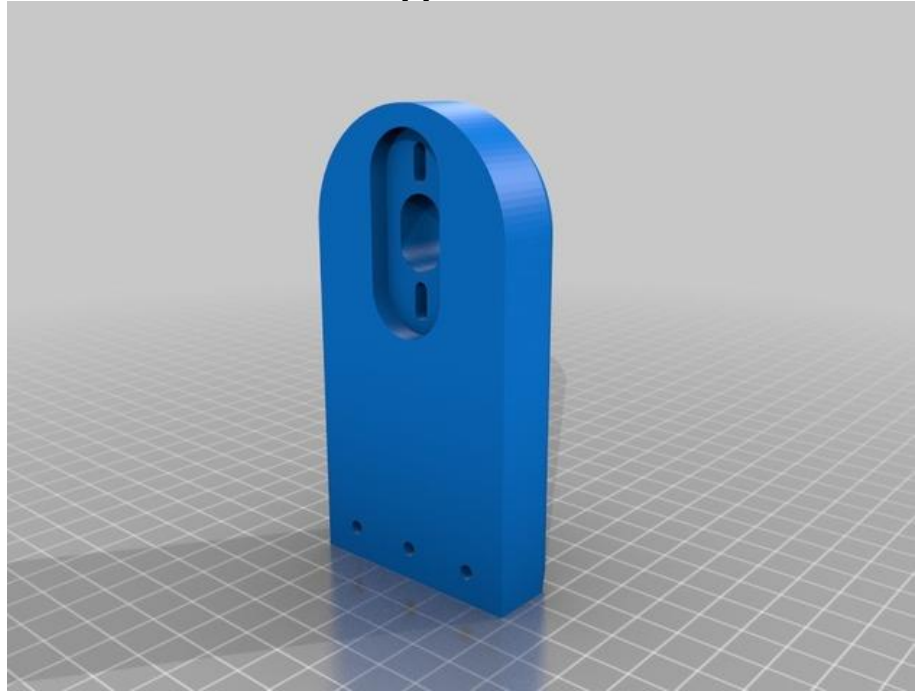

**Fig : 4.10**

### 4.1.10 Upper Arm Right

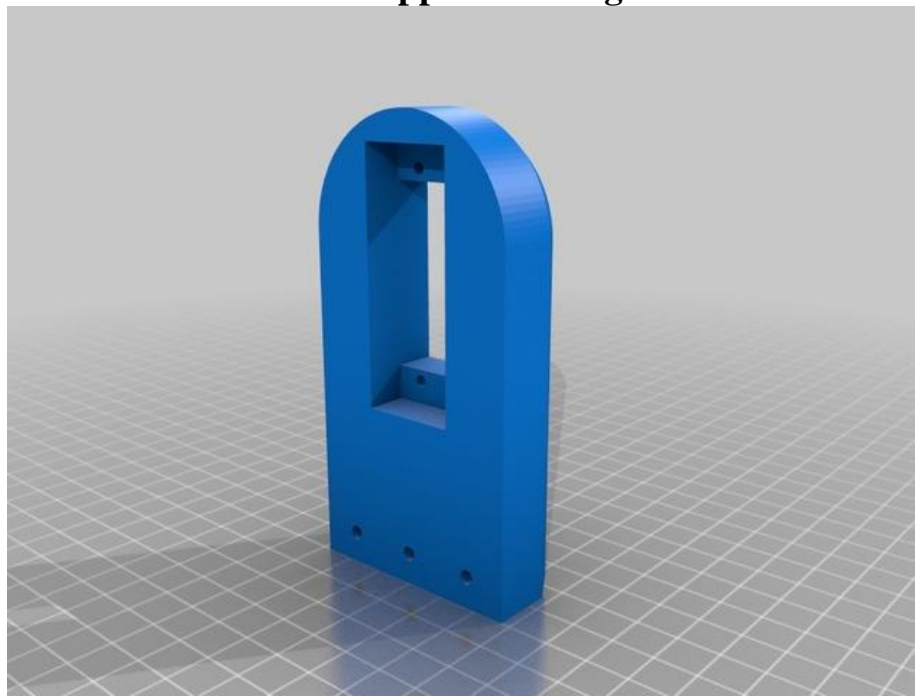

**Fig : 4.11**

### 4.1.11 Upper Arm Servo Mount

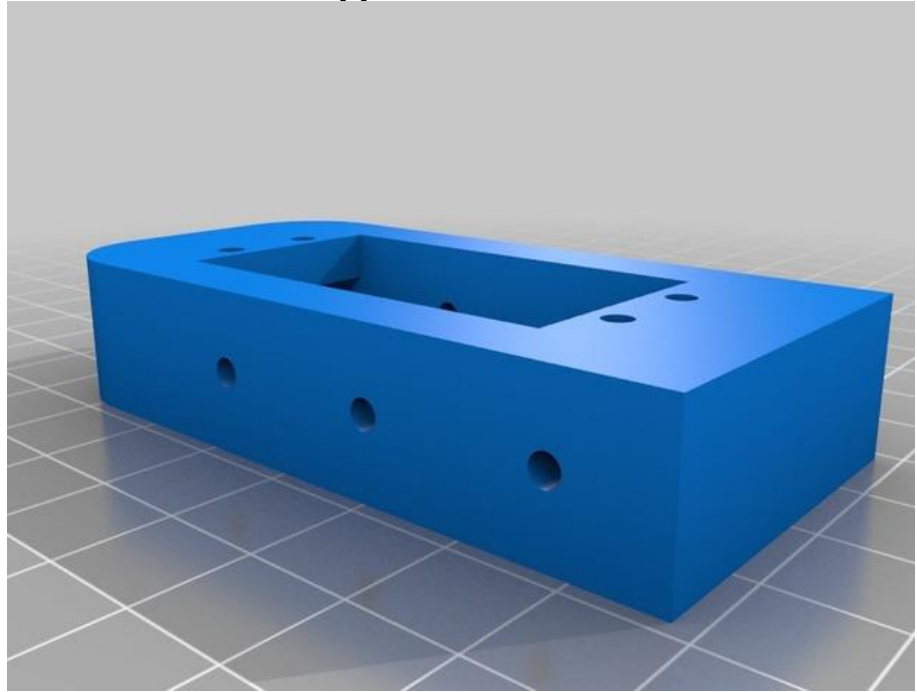

**Fig : 4.12**

## 4.2 Electrical Circuits

### 4.2.1 Codes

For **Gyroscopic Stability**

```cpp
#include <Wire.h>

const float accScale= 9.81f/ 16384;

const float rotScale= 1.f / 131;

class Gyro

{

public:

uint8_t address;

float positionA[3];

float rotationV[3];

int correctionR[3];

float temperature;

int oldTime;

int gscale, ascale;

Gyro(int ascale = 0, int gscale = 0, uint8_t address = 0x68)

{

this->address = address;
```

```
//wake up call

this->ascale = ascale;

this->gscale = gscale;

correctionR[0] = correctionR[1] = correctionR[2] = 0;

oldTime = 0;

}

void begin()

{

Wire.begin();

wakeUp();

setScale();

}

void wakeUp()

{

Wire.beginTransmission(address);

Wire.write(0x6B);

Wire.write(0);

Wire.endTransmission(true);

}

void setScale()
```

```
    {
        Wire.beginTransmission(address);
        Wire.write(0x1B);
        Wire.write(gscale << 3);
        Wire.endTransmission(true);
        Wire.beginTransmission(address);
        Wire.write(0x1C);
        Wire.write(ascale << 3);
        Wire.endTransmission(true);
    }
    inline signed short readShort() const
    {
        return (Wire.read() << 8) | Wire.read();
    }
    void calculateCorrection(int samples = 100)
    {
        long dr[3] = {0, 0, 0};
        for(int i = 0; i < samples; i++)
        {
            Wire.beginTransmission(address);
```

```
        Wire.write(0x3B);

        Wire.endTransmission(false);

        Wire.requestFrom((uint8_t)address, (uint8_t)14, (uint8_t)true);

        positionA[0] = readShort() * accScale * (1 << ascale);

        positionA[1] = readShort() * accScale * (1 << ascale);

        positionA[2] = readShort() * accScale * (1 << ascale);

        temperature = readShort() / 340.f + 36.53f;

        dr[0] += readShort();

        dr[1] += readShort();

        dr[2] += readShort();

        delay(10);

    }

    correctionR[0] = dr[0] / samples;

    correctionR[1] = dr[0] / samples;

    correctionR[2] = dr[0] / samples;

}

int poll()

{

    Wire.beginTransmission(address);

    Wire.write(0x3B);
```

```
Wire.endTransmission(false);

Wire.requestFrom((uint8_t)address, (uint8_t)14, (uint8_t)true);

positionA[0] = readShort() * accScale * (1 << ascale);

positionA[1] = readShort() * accScale * (1 << ascale);

positionA[2] = readShort() * accScale * (1 << ascale);

temperature = readShort() / 340.f + 36.53f;

rotationV[0] = (readShort() - correctionR[0]) * rotScale * (1 << gscale);

rotationV[1] = (readShort() - correctionR[1]) * rotScale * (1 << gscale);

rotationV[2] = (readShort() - correctionR[2]) * rotScale * (1 << gscale);

}

};
```

For **Matrix**

#pragma once

#include <math.h>

class Vector

{

public:

float v[4];

Vector(float x = 0, float y = 0, float z = 0, float w = 1)

{

v[0] = x;

v[1] = y;

v[2] = z;

v[3] = w;

}

Vector operator*(float s) const

{

return Vector(v[0] * s, v[1] * s, v[2] * s, 1);

}

Vector operator+(const Vector &v2) const

```cpp
{
    return Vector(v[0] + v2.v[0], v[1] + v2.v[1], v[2] + v2.v[2], 1);
}

Vector operator-(const Vector &v2) const
{
    return Vector(v[0] - v2.v[0], v[1] - v2.v[1], v[2] - v2.v[2], 1);
}

Vector &operator*=(float s)
{
    *this = *this * s;
    return *this;
}

float &operator[](int i)
{
    return v[i];
}

float length()
{
    return sqrt(v[0] * v[0] + v[1] * v[1] + v[2] * v[2]);
}
```

```cpp
void normalize()
{
    float l2 = v[0] * v[0] + v[1] * v[1] + v[2] * v[2];
    if(!l2) return;
    float rl = 1 / sqrt(l2);
    *this *= rl;
}

float dot(const Vector &v2) const
{
    return v[0] * v2.v[0] + v[1] * v2.v[1] + v[2] * v2.v[2];
}

Vector cross(const Vector &v2) const
{
    return Vector(v[1] * v2.v[2] - v[2] * v2.v[1], v[2] * v2.v[0] - v[0] * v2.v[2], v[0] * v2.v[1] - v[1] * v2.v[0]);
}

Vector project(const Vector &v2) const
{
    Vector nv2 = v2;
    nv2.normalize();
    float s = dot(nv2);
```

```cpp
        return nv2 * s;

    }

};

class Matrix

{

public:

    float m[4][4];

    Matrix()

        :Matrix(1, 0, 0, 0,  0, 1, 0 ,0,  0, 0, 1, 0, 0, 0, 0, 1)

    {

    }

    Matrix(float m00, float m01, float m02, float m03,

        float m10, float m11, float m12, float m13,

        float m20, float m21, float m22, float m23,

        float m30, float m31, float m32, float m33)

    {

        m[0][0] = m00; m[0][1] = m01; m[0][2] = m02; m[0][3] = m03;

        m[1][0] = m10; m[1][1] = m11; m[1][2] = m12; m[1][3] = m13;

        m[2][0] = m20; m[2][1] = m21; m[2][2] = m22; m[2][3] = m23;

        m[3][0] = m30; m[3][1] = m31; m[3][2] = m32; m[3][3] = m33;
```

```cpp
}

static Matrix identity()

{

return Matrix(1, 0, 0, 0,  0, 1, 0 ,0,  0, 0, 1, 0, 0, 0, 0, 1);

}

static Matrix scaling(float s)

{

return Matrix(s, 0, 0, 0,  0, s, 0 ,0,  0, 0, s, 0, 0, 0, 0, 1);

}

static Matrix scaling(float u, float v, float w)

{

return Matrix(u, 0, 0, 0,  0, v, 0 ,0,  0, 0, w, 0, 0, 0, 0, 1);

}

static Matrix translation(float x, float y, float z)

{

return Matrix(1, 0, 0, x,  0, 1, 0, y,  0, 0, 1, z, 0, 0, 0, 1);

}

static Matrix rotation(float a, float x, float y, float z)

{

float cosa = cos(a);
```

```
        float rcosa = 1 - cosa;
        float sina = sin(a);
        return Matrix(
            x * x * rcosa + cosa,    x * y * rcosa - z * sina, x * z * rcosa + y * sina, 0,
            y * x * rcosa + z * sina, y * y * rcosa + cosa,    y * z * rcosa - x * sina, 0,
            z * x * rcosa - y * sina, z * y * rcosa + x * sina, z * z * rcosa + cosa, 0,
            0, 0, 0, 1
        );
    }
    static Matrix perspective(float fov, float near, float far)
    {
        float scale = tan(fov * 0.5 * M_PI / 180);
        return Matrix(
            scale, 0, 0, 0,
            0, scale, 0, 0,
            0, 0, -far * near / (far - near), 0,
            0, 0, -1, 0
        );
```

```cpp
}

Vector operator *(const Vector &v)
{
return Vector(
v.v[0] * m[0][0] + v.v[1] * m[0][1] + v.v[2] * m[0][2] + m[0][3],
v.v[0] * m[1][0] + v.v[1] * m[1][1] + v.v[2] * m[1][2] + m[1][3],
v.v[0] * m[2][0] + v.v[1] * m[2][1] + v.v[2] * m[2][2] + m[2][3],
v.v[0] * m[3][0] + v.v[1] * m[3][1] + v.v[2] * m[3][2] + m[3][3]
);
}

Matrix operator *(const Matrix &m2)
{
Matrix mr;
for(int y = 0; y < 4; y++)
for(int x = 0; x < 4; x++)
mr.m[y][x] = m[y][0] * m2.m[0][x] + m[y][1] * m2.m[1][x] + m[y][2] * m2.m[2][x] + m[y][3] * m2.m[3][x];
return mr;
}

Matrix &operator *=(const Matrix &m2)
{
```

```
    *this = *this * m2;

    return *this;

    }

};
```

**FINAL CODE**

```
#include <Wire.h>

//PWM library

#include <Adafruit_PWMServoDriver.h>

//compass sensor library

#include <Adafruit_Sensor.h>

#include <Adafruit_LSM303_U.h>

//matrix and vector implementation

#include "Matrix.h"

//MPU gyro sensor implementation

#include "Gyro.h"

const int servoCount = 6;

//constants for pulse widths at 12 bit PWM

const int servoPulseMin = 128;

const int servoPulseMax = 640;

const int servoPulseWidth = servoPulseMax - servoPulseMin;

//analog normatization values
```

```
int analogMin = 0;

const int analogMax = 1023;

//the from 0 to 0.99999. the higher the smoother the movements but also more delayed

const float attenuation = 0.8;

//reference compass vector

Vector compass0;

//servo ranges [min, max] from 0 to 1. This prevents that the arm hits itself and the servos overheat

float servoRange[servoCount][2] = {

{0.2, 0.9},

{0.35, 0.95},

{0.0, 1},

{0.25, 1},

{0, 0.75},

{0.15, 0.4}};

Vector sensors[4];  //vectors of the sensors

float analog = 0; //last analog reading of the hall effect
```

```cpp
float angles[servoCount] = {-1};  //angle values of the joints

float servoValue[servoCount] = {0.5, 0.4, 0.2, 0.25, 0.0, 0.2}; //values the servos will be set to [0; 1]

int reversedRotation[servoCount] = {0, 1, 0, 1, 0, 0};  //needed to reverse rotation for servos that are flipped

float offsetValue[servoCount] = {0.5, 0.4, 0.2, 0.25, 0.0, 0.2}; //offset for the angle 0

//sensor class initializations

Gyro accel1(0, 1, 0x68);

Gyro accel2(0, 1, 0x69);

Adafruit_LSM303_Accel_Unified accel3 = Adafruit_LSM303_Accel_Unified(54321);

Adafruit_LSM303_Mag_Unified mag1 = Adafruit_LSM303_Mag_Unified(12345);

//pwm driver

Adafruit_PWMServoDriver pwm = Adafruit_PWMServoDriver(0x40);

///reads all sensor values and normalizes all except for the magnetometer

void readSensors()
```

```
{
    //read the sensor vlaues
    accel1.poll();
    accel2.poll();
    sensors_event_t accel3e;
    accel3.getEvent(&accel3e);
    sensors_event_t mag1e;
    mag1.getEvent(&mag1e);

    //set the vectors and normalize
    sensors[0] = Vector(accel1.positionA[0], accel1.positionA[1], accel1.positionA[2]);
    sensors[0].normalize();

    sensors[1] = Vector(accel2.positionA[0], accel2.positionA[1], accel2.positionA[2]);
    sensors[1].normalize();

    sensors[2] = Vector(accel3e.acceleration.x, accel3e.acceleration.y, accel3e.acceleration.z);
    sensors[2].normalize();
```

```cpp
sensors[3] = Vector(mag1e.magnetic.x, mag1e.magnetic.y, mag1e.magnetic.z);

//sensors[3].normalize();

//scale the analog reading to [0; 1]
analog = float(analogRead(A0) - analogMin) / (analogMax - analogMin);
}

///initial calibration, reference values for the gripper and the yaw are read
void calibrate()
{
//get minimal value when gipper is open
analogMin = analogRead(A0);
//read the sensors
readSensors();
//calculate the absolute compass vector
float a1 = -atan2(-sensors[1][1], -sensors[1][0]);
Matrix r0 = Matrix::rotation(-a1, 0, 1, 0);
float a2 = atan2(sensors[2][1], sensors[2][2]);
```

```
    Matrix r1 = Matrix::rotation(angles[4], 1, 0, 0);

    Vector v = r0 * r1 * sensors[3];

    v[2] = 0;

    v.normalize();

    compass0 = v;

}

void setup()

{

//setup serial

Serial.begin(115200);

Serial.println("Setup");

//enable gyro sensors

accel1.begin();

accel2.begin();

if(!accel3.begin())

Serial.println("LSM303 accelerometer not detected");

//enable magnetometer

if(!mag1.begin())

Serial.println("LSM303 magnetometer not detected");
```

```cpp
//enable pwm board and set the frequency to 50Hz

pwm.begin();

pwm.setPWMFreq(50);

for(int i = 0; i < servoCount; i++)

pwm.setPWM(i, 4096, 0);

//get initial readings

calibrate();

}

///calculate the servo values form the angles

void setServoAngle(int servo, float angle)

{

//convert to [0; 1] scale and clip to min max range

float v = max(servoRange[servo][0], min(servoRange[servo][1], angle / float(M_PI) + offsetValue[servo]));

//attenuate if not forst values

if(servoValue[servo] >= 0)

servoValue[servo] = servoValue[servo] * attenuation + v * (1 - attenuation);

else

servoValue[servo] = v;
```

}

///writes the servo values to the servos

void updateServos()

{

for(int i = 0; i < servoCount; i++)

if(servoValue[i] < 0)

pwm.setPWM(i, 4096, 0);

else

pwm.setPWM(i, 0, int(servoPulseMin + (reversedRotation[i] ? 1 - servoValue[i]: servoValue[i]) * servoPulseWidth));

}

///turn off a servo

void disableServo(int servo)

{

servoValue[servo] = -1;

}

///outputs all measured sensor values

void printSensors()

```
{
    for(int i = 0; i < 4; i++)
    {
        for(int j = 0; j < 3; j++)
        {
            Serial.print(sensors[i][j]); Serial.print(' ');
        }
        Serial.print("| ");
    }
    Serial.print(analog);
    Serial.print("   ");
}

///calculates the angles from the sensor values
void calculateAngles()
{
    //shoulder roll
    angles[1] = -atan2(-sensors[0][2], -sensors[0][1]);
    //shoulder pitch
    angles[2] = atan2(-sensors[0][0], -sensors[0][1]);
```

```cpp
float elbowPitch = atan2(-sensors[1][0], -sensors[1][1]);

//setting relative pitch to shoulder

angles[3] = elbowPitch - angles[2];

//wrist roll

angles[4] = atan2(sensors[2][1], sensors[2][2]);

//gripper depending on analog value. We take the the square toot to get a linear movement. (point sources attenuate quadratically)

angles[5] = (sqrt(max(analog, 0.f)) * (servoRange[5][1] - servoRange[5][0])) * M_PI;

//calculating reverse tansformation for lower arm pitch

float a1 = -atan2(-sensors[1][1], -sensors[1][0]);

Matrix r0 = Matrix::rotation(-a1, 0, 1, 0);

//calculating reverse transformation of wrist roll

float a2 = atan2(sensors[2][1], sensors[2][2]);

Matrix r1 = Matrix::rotation(angles[4], 1, 0, 0);

//transforming the magnetic field vector back

Vector v = r0 * r1 * sensors[3];

//switching to flat earth compass
```

```
    v[2] = 0;

    v.normalize();

    //calculating angle between startup and current compass value

    Vector a = v.cross(compass0);

    angles[0] = -asin(a[2]);

    //setting the values

    setServoAngle(0, angles[0]);

    setServoAngle(1, angles[1]);

    setServoAngle(2, angles[2]);

    setServoAngle(3, angles[3]);

    setServoAngle(4, angles[4]);

    setServoAngle(5, angles[5]);

}

void printAngles()

{

    for(int i = 0; i < servoCount; i++)

    {

        Serial.print(angles[i]); Serial.print(' ');
```

```
   }
   Serial.println();
}

///main loop
void loop()
{
   static int lastServoUpdate = 0;
   //get current time
   int t = millis();
   static int time = 0;
   int dt = t - time;
   if (time == 0) dt = 0;
   time = t;

   //read sensors, calculate angles and set servos
   //servo PWM cant be updated more often than every 20ms
   if(time - lastServoUpdate > 20)
   {
      lastServoUpdate = time;
```

readSensors();

printSensors();

calculateAngles();

printAngles();

updateServos();

}

}

### 4.2.2 Circuit Logic

The microcontroller uses I$^2$C to talk to the most of the components. The first GY-521 sensor connects ground to the address pin and the second GY-521 connects VCC to get different addresses. The GY-511 does not have a special configuration. The Hall Effect sensor is connected to ground VCC and to the analogue pin of the microcontroller. The PWM board also uses I$^2$C but requires an additional power source to deliver enough current to the servo motors.

After testing, we would have transferred the flimsy bread board set up to perfboard (It is a material for prototyping electronic circuits, also known as DOT PCB. It is a thin, rigid sheet with holes pre-drilled and ringed by round or square copper pads at standard intervals across a grid, usually a square grid of 2.54 mm spacing) with the help of some pin headers.

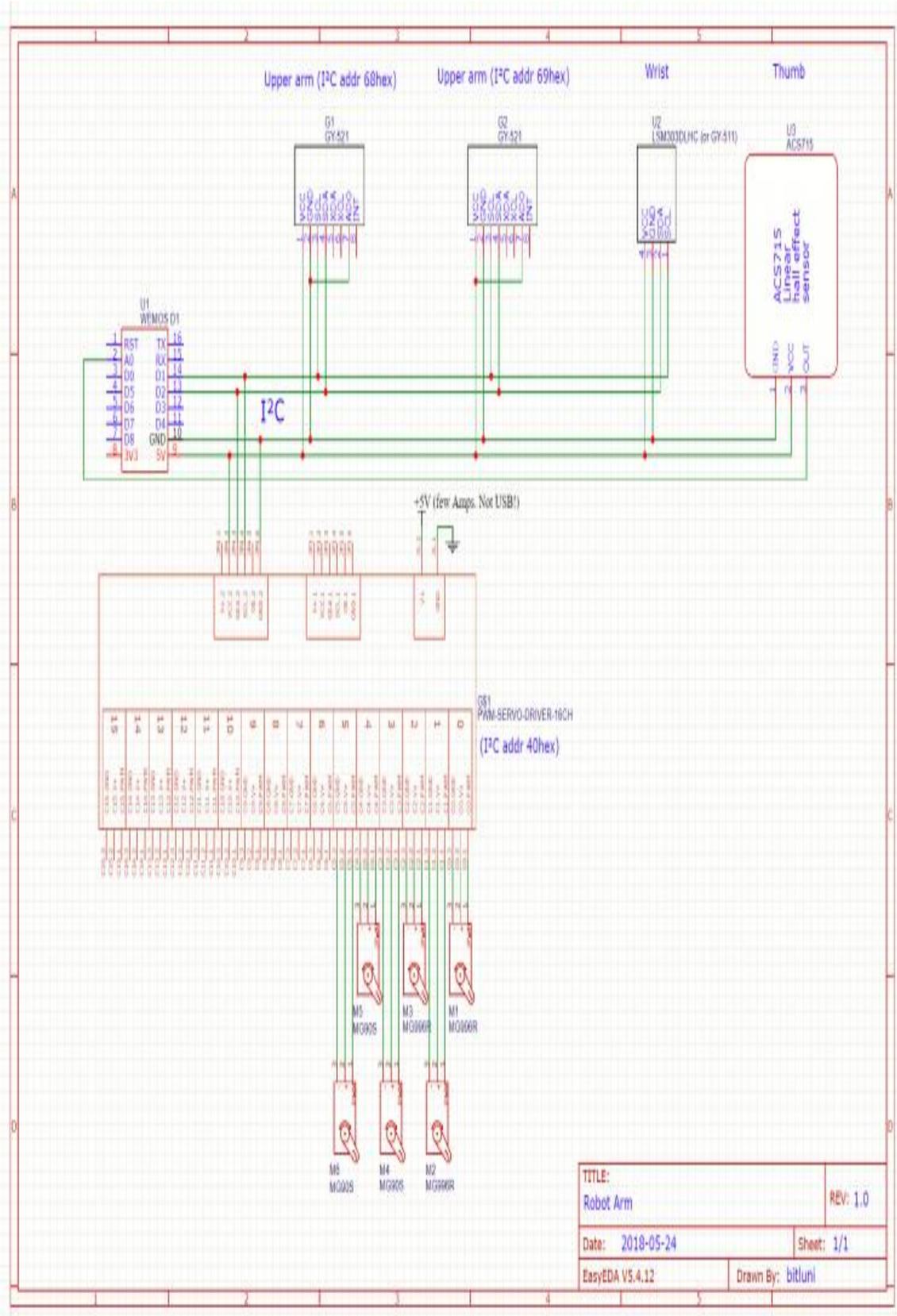

**Fig 4.13**

## 4.3 Assembly

The work of Assembly of parts was done on Blender and AutoCAD. While the work of electric circuits was done on Arduino IDE, GitHub and MATLAB. The Arduino IDE is a cross-platform application that is written in functions from C and C++. It is used to write and upload programs to Arduino compatible boards, but also, with the help of 3rd party cores, other vendor development boards. MATLAB is a multi-paradigm numerical computing environment and proprietary programming language. MATLAB allows matrix manipulations, plotting of functions and data, implementation of algorithms, creation of user interfaces, and interfacing with programs written in other languages.

### 4.3.1 Final Assembly View in Blender Software

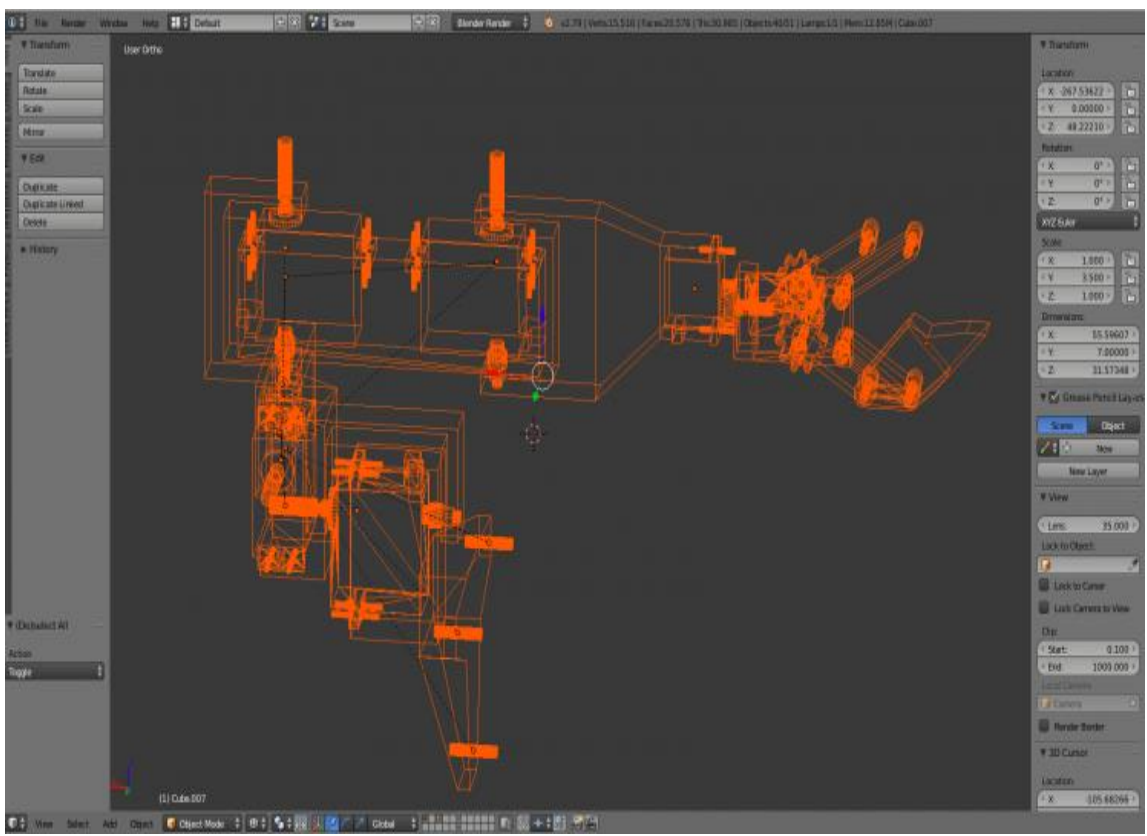

**Fig 4.14**

### 4.3.2 Rendered View of Assembly

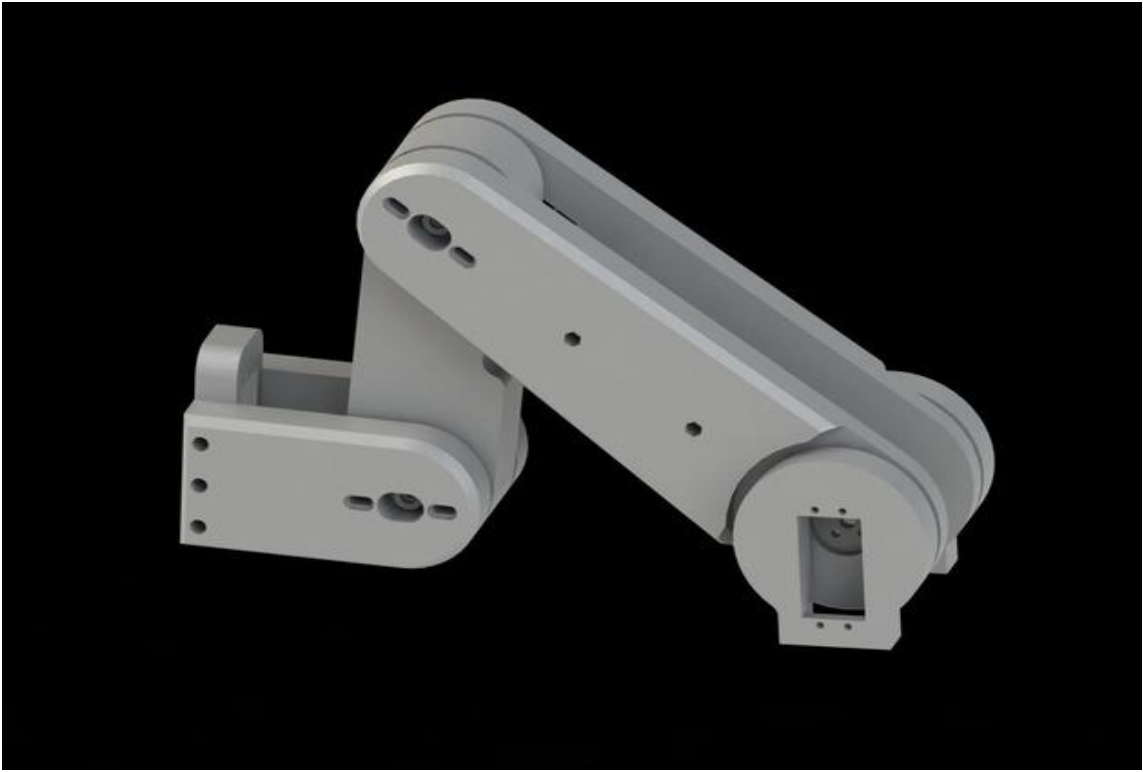

**Fig 4.15**

### 4.3.3 Expanded View of the Rendered Assembly

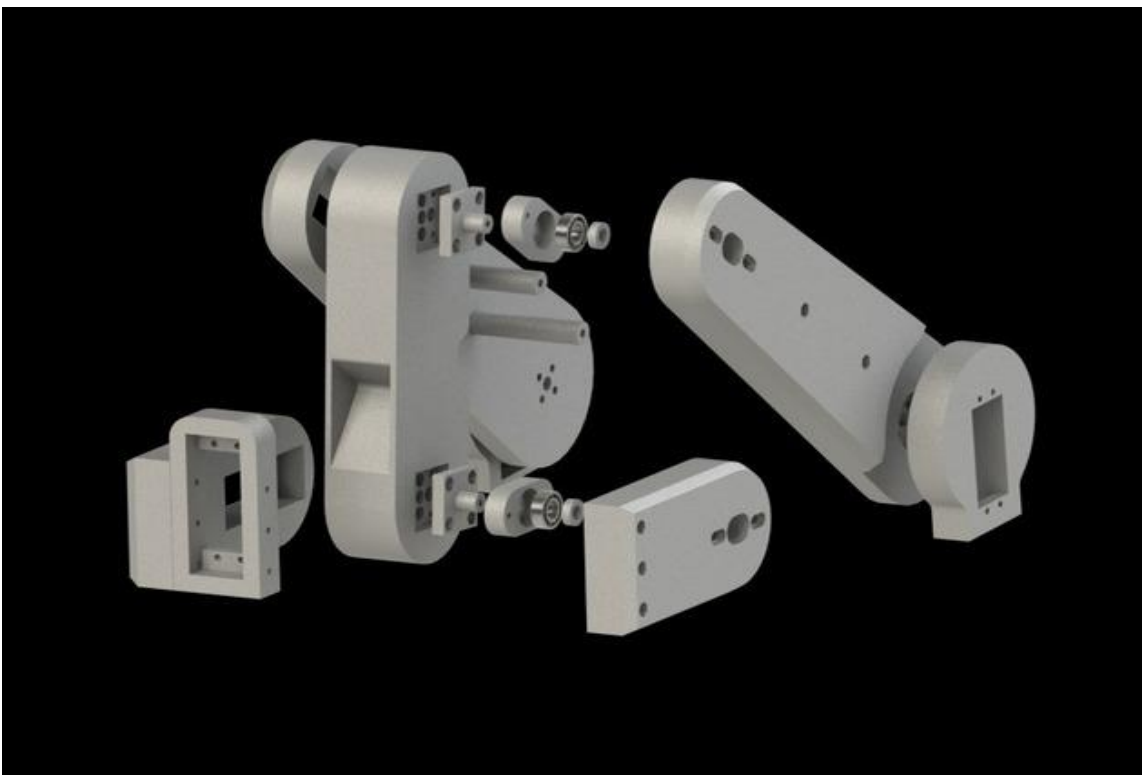

**Fig 4.16**

## 4.4 Project Process Overview and Flowchart

The project basically involves a set of parts which have to be 3D printed. Also, here I am using 3D printing as we can achieve high precision at a lower cost. Another benefit along with lower cost is the reduction in weight of the hand as a whole. The CAD files of the same are attached in the link.

The arm is designed to have all components fit in itself. After the parts will arrive from the printing, we will integrate & assemble the electrical components and will make the required adjustments if needed.

After integrating all the components together, we would wire all the motors to the Arduino bread module. We have not yet used PCB, instead we are using Bread Module to run the testing codes on our first prototype as we might need to change the wiring.

The next step is to add the codes to the Arduino library. The codes are designed in a way to take reading from the accelerometers in the artificial arm and calculate the degree by which the motors have to rotate their shaft.

Next is to calibrate the motors with the Arduino and check if the accuracy and precision are up to the mark. Then the whole arm is booted up along with the external system and the arm starts imitating the movements of your hands.

# Chapter 5

# CONCLUSION AND FUTURE SCOPE

The main people who will be benefited from the project will be the physically challenged population. Something similar has been developed certain years ago but the mechanism we are using hasn't been used yet and would provide an artificial limb with similar working but at a very lower price.

Final product could also be used by industries (other than space agencies medical industry) with minimal changes like memory system and different gripping mechanism.

Plus the labour cost has increased significantly over the past decade. In order to reduce the cost as well as increasing the efficiency and reducing the human error, a product needs to be developed. The final product of this project will help you achieve the above stated goal.

Till date, the robotic arms made have been using the normal sensors that could replicate the motion of a human arm but we are using a different mechanism in which we will be working on an algorithm to replicate the motion of human arm which first captures the motion, then with the help of algorithm, it will measure the work field as a 3 axis cartesian plane. It will then locate the final point in the defined space and then find the 6 different angles that each motor will have to adjust, in order for the tool to reach the coordinates.

# REFFERENCES